\documentclass[tinyml]{acmart}

\AtBeginDocument{%
  \providecommand\BibTeX{{%
    \normalfont B\kern-0.5em{\scshape i\kern-0.25em b}\kern-0.8em\TeX}}}

\setcopyright{rightsretained}
\copyrightyear{2023}
\acmYear{2023}

\usepackage{glossaries}
\usepackage{listings}

\newcommand{\rotmulticell}[2][c]{%
  \rotatebox{90}{\begin{tabular}[#1]{@{}c@{}}#2 \end{tabular}}}

\lstdefinestyle{myStyle}{
  language=Python,
  numbers=left,
  stepnumber=1,
  tabsize=4,
  showspaces=false,
  showstringspaces=false,
  basicstyle=\ttfamily \scriptsize,
  xleftmargin=5.0ex,
  morekeywords={func, for, in, if, return, while, True, break, endfor, endwhile, then, do, False}
}

\lstdefinestyle{myStyleText}{
  language=C,
  numbers=left,
  stepnumber=1,
  tabsize=4,
  showspaces=false,
  showstringspaces=false,
  basicstyle=\scriptsize,
  xleftmargin=5.0ex,
  mathescape=true,
  basewidth=0.5em,
  literate=
    {:=}{$\leftarrow{}$}{1}
    {->}{$\rightarrow{}$}{1},
  morekeywords={func, for, in, if, return, while, True, break, endfor, endwhile, then, do, False}
}

\newacronym{cnn}{CNN}{Convolutional Neural Network}
\newacronym{dnn}{DNN}{Deep Neural Network}
\newacronym{iot}{IoT}{Internet of Things}
\newacronym{ilp}{ILP}{Integer Linear Programming}
\newacronym{mip}{MIP}{Mixed Integer Programming}
\newacronym{milp}{MILP}{Mixed Integer Linear Program}
\newacronym{ir}{IR}{Intermediate Representation}
\newacronym{nas}{NAS}{Network Architecture Search}
\newacronym{fdt}{FDT}{Fused Depthwise Tiling}
\newacronym{ffmt}{FFMT}{Fused Feature Map Tiling}

\begin{document}

\title{Fused Depthwise Tiling for Memory Optimization in TinyML Deep Neural Network Inference}

\author{Rafael Stahl, Daniel Mueller-Gritschneder, Ulf Schlichtmann}
\email{{r.stahl, daniel.mueller, ulf.schlichtmann}@tum.de}
\affiliation{%
  \institution{Technical University of Munich}
  \city{Munich}
  \country{Germany}
}

\renewcommand{\shortauthors}{Rafael Stahl, et al.}

\begin{abstract}

Memory optimization for deep neural network (DNN) inference gains high relevance with the emergence of TinyML, which refers to the deployment of DNN inference tasks on tiny, low-power microcontrollers.
Applications such as audio keyword detection or radar-based gesture recognition are heavily constrained by the limited memory on such tiny devices because DNN inference requires large intermediate run-time buffers to store activations and other intermediate data, which leads to high memory usage.
In this paper, we propose a new Fused Depthwise Tiling (FDT) method for the memory optimization of DNNs, which, compared to existing tiling methods, reduces memory usage without inducing any run time overhead.
FDT applies to a larger variety of network layers than existing tiling methods that focus on convolutions.
It improves TinyML memory optimization significantly by reducing memory of models where this was not possible before and additionally providing alternative design points for models that show high run time overhead with existing methods.
In order to identify the best tiling configuration, an end-to-end flow with a new path discovery method is proposed, which applies FDT and existing tiling methods in a fully automated way, including the scheduling of the operations and planning of the layout of buffers in memory.
Out of seven evaluated models, FDT achieved significant memory reduction for two models by 76.2\% and 18.1\% where existing tiling methods could not be applied.
Two other models showed a significant run time overhead with existing methods and FDT provided alternative design points with no overhead but reduced memory savings.

\end{abstract}

\keywords{neural networks, embedded software}

\maketitle

\section{Introduction}

Edge machine learning applications offer superior possibilities over cloud computing approaches in terms of communication demand, latency and data privacy.
Edge devices have a wide range of computation classes.
It was shown that for certain machine learning workloads, the inference of \glspl{dnn} can be performed even on tiny, low-power microcontroller-type devices.
The solutions that enable \gls{dnn} inference on microcontrollers are known as \textit{TinyML} or \textit{Extreme Edge AI} and were successfully applied to applications such as keyword spotting, visual wake-up, anomaly detection or radar-based gesture recognition.
A central challenge that TinyML deployment faces is the narrowly constrained embedded memory that may only offer several hundred kB of SRAM.
Therefore, TinyML solutions include methods to reduce the memory usage of any given \gls{dnn} inference task, such as quantization, pruning and \gls{nas}.
All these methods have in common that they change \gls{dnn} parameters and, therefore, the \gls{dnn}'s behavior and inference results.
One method to reduce memory usage without changing any \gls{dnn} behavior is fused tiling.
If the lifetimes of two intermediate tensor buffers do not overlap, their storage buffers may overlap, allowing to reduce the overall memory demand.
The size of intermediate buffers can be reduced by mutating the DNN graph with \textit{tiling} and their lifetimes can be decoupled by \textit{fusing} multiple consecutive operations.


The main contribution of this paper is the introduction of \gls{fdt} for the memory optimization of DNNs.
By tiling depthwise (i.e., by channels for convolutions), new tiling opportunities are enabled that reduce peak memory usage without any run time overheads that would be induced by existing tiling methods.
Even though convolutions and dense operations require all inputs for each output, two of them can be fused to tile an intermediate buffer.
Additionally, FDT can be applied to more layer types than existing methods that focus solely on convolutions, such that a wider range of models can be tiled.
The new FDT tiling method overall improves TinyML memory optimization significantly, not by replacing existing tiling methods, but expanding the tiling design space in combination with existing methods.
To explore this expanded tiling design space, we provide an end-to-end deployment flow that automatically determines where and how to apply fused tiling optimally on any given \gls{dnn}.
Exploitation of the tiled graphs for memory reduction additionally requires a suitable memory-aware scheduling of operations and memory buffer layout planning. Hence, these two steps are also automated and efficiently implemented to conduct a fast exploration.
To quickly find optimized tiling opportunities, we also run a process called path discovery that analyzes the DNN graph and explores possible tiling configurations.
In summary, our contributions are as follows.
\begin{enumerate}
    \item The tiling method \gls{fdt} applied for the memory optimization of \glspl{dnn} to expand the design space by reducing memory further or eliminating run time overheads.
    \item An automated exploration with a new block-based path discovery to find suitable tiling configurations, a memory-aware scheduling and optimal memory layout planning.
\end{enumerate}

In a sample of seven models that benefit from fused tiling, FDT achieved significant memory reduction for two models by 76.2\% and 18.1\% where existing tiling methods could not be applied.
Two other models showed a significant run time overhead with existing methods and FDT provided an alternative design point with no overhead but reduced memory savings.

\section{Related Work}

Inference on resource-constrained devices can be tackled in a number of ways.
Offloading computation to other infrastructure such as the cloud is widely used, but introduces challenges of high bandwidth and energy requirements for data transfer, network latency and privacy concerns~\cite{kang2017neurosurgeon}.
Orthogonal methods to fused tiling are quantization~\cite{gholami2021survey}, pruning~\cite{guo2020channel} and \gls{nas}~\cite{lin2020mcunet,banbury2021micronets}.

\begin{figure}[tb]
    \centerline{\includegraphics[width=0.40\textwidth]{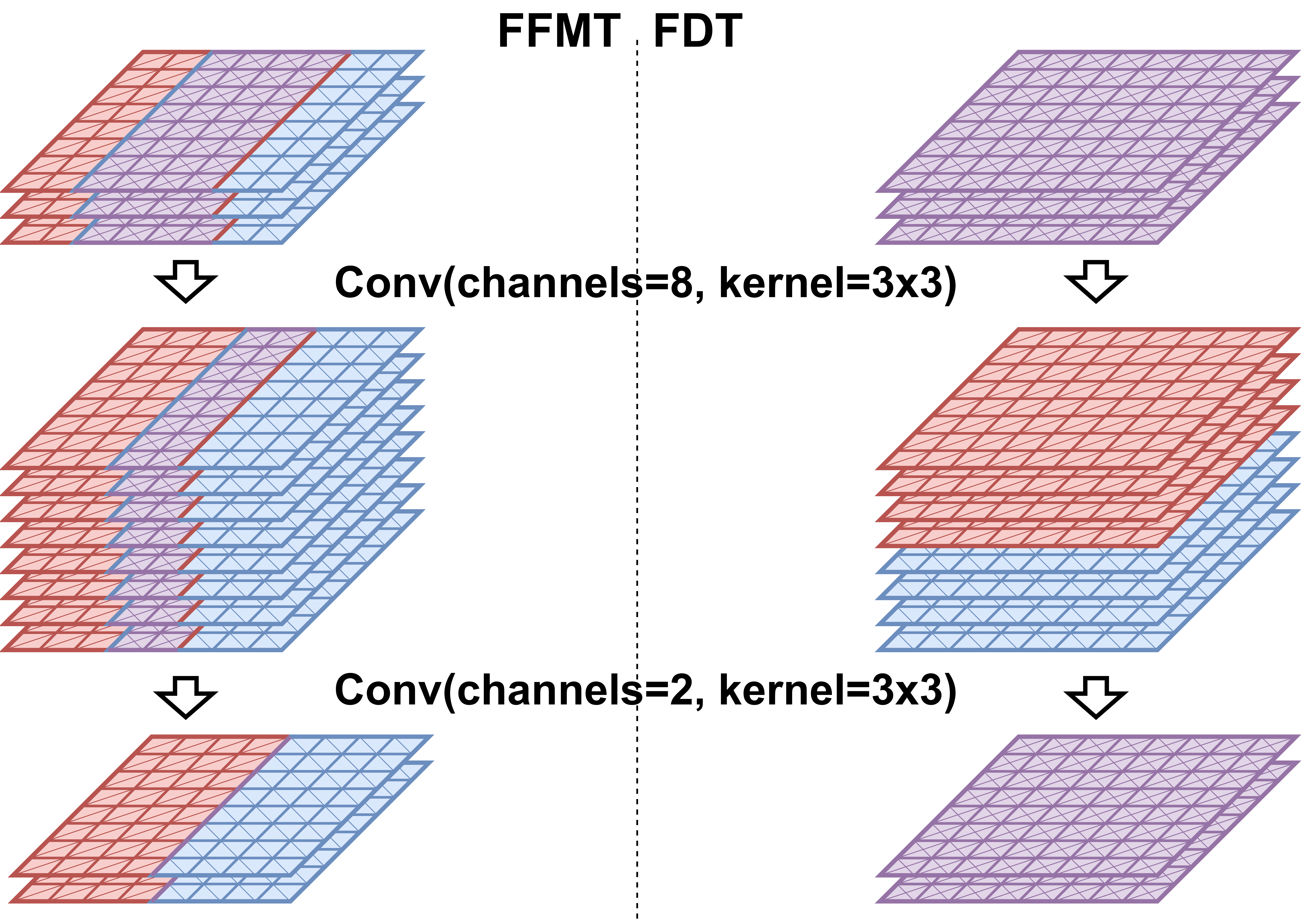}}
    \caption{FFMT compared to FDT.}
    \label{fig:ffmt_vs_fdt}
\end{figure}

Tiling is the splitting of \gls{dnn} graph operations such that individual partitions can be computed independently of each other.
It is used primarily within a single \gls{dnn} operation to accelerate execution~\cite{lin2020mcunet}.
Another application of tiling is the partitioning of \glspl{dnn} so that they can be run distributed over several devices~\cite{zhao2018deepthings} or can be offloaded to one or more accelerators~\cite{alwani2016fused}.
A novel aspect of these works are their use of fusing to keep consecutive tiled partitions independent of each other.
The basic principle is summarized in Fig.~\ref{fig:ffmt_vs_fdt} on the left and will be referred to as \gls{ffmt}.
The figure shows two consecutive convolutional operations as part of a \gls{dnn}.
The three sets of feature maps are the input of the first operation, an intermediate buffer and the output of the second operation.
Since the intermediate buffer is larger than the input and output, tiling it could reduce memory requirements.
\Gls{ffmt} does this by splitting all feature maps of the intermediate tensor buffer into partitions.
Convolution operations have spatial locality, which allows to produce output feature maps from split inputs mostly independently.
However, convolution kernels larger than 1x1 cause an overlap in the input partitions that accumulates additively over all tiled operations.
Two partitions are shown in Fig.~\ref{fig:ffmt_vs_fdt} in different colors/patterns, and their overlap caused by 3x3 convolutions is marked in purple/crossed.
\Gls{ffmt} was first employed for reducing peak memory usage in~\cite{cipolletta2021dataflow}, but their path discovery requires partially manual user effort.
Other works that use \gls{ffmt} with automated path discovery are~\cite{arm2021cascading,minakova2020buffer,minakova2022memory,lin2021mcunetv2,colleman2021high}.
\Gls{ffmt} along with tiling in the depthwise dimension for single layers without operator fusion was explored in~\cite{burrello2021dory,mei2021zigzag}.
Based on these, the work in~\cite{mei2022defines} identifies the full \gls{ffmt} search space of loop scheduling in a memory hierarchy and adds a new cost model.
\cite{mei2022defines}~further states: "in most convolution layers all input channels are required to calculate any output feature, which makes cross-layer tiling across the channel dimensions impossible", which we challenge with this work.
Our distinction from existing work is summarized in Table~\ref{tab:tiling_methods}.
Our work is the first to exploit \gls{fdt} for optimization of working memory (RAM) in \gls{dnn} inference while also keeping the path discovery fully automated.
\Gls{fdt} will be explained in full detail in the following section.

\begin{table}[tb]
    \caption{Comparison of Tiling Methods}
    \centering
    \scalebox{0.76}{
    \begin{tabular}{|l|cc|}
        \hline
        \textbf{Work}   & \textbf{FFMT} & \textbf{FDT}  \\
        \hline
        Distributed Inference~\cite{zhao2018deepthings}  & RAM reduction & -  \\
        Full Distributed Inference~\cite{stahl2021deeperthings}  & RAM reduction & ROM reduction  \\
        Partly Manual Tiling~\cite{cipolletta2021dataflow,arm2021cascading} & RAM reduction & -  \\
        Automated Tiling~\cite{minakova2020buffer,minakova2022memory,lin2021mcunetv2,colleman2021high,burrello2021dory,mei2021zigzag,mei2022defines} & RAM reduction & -  \\
        \hline
        Our Automated Tiling  & RAM reduction & RAM reduction  \\
        \hline
    \end{tabular}}
    \label{tab:tiling_methods}
\end{table}

\section{Fused Depthwise Tiling (FDT)}
\label{sec:fdt}

\begin{figure}[tb]
    \centerline{\includegraphics[width=0.33\textwidth]{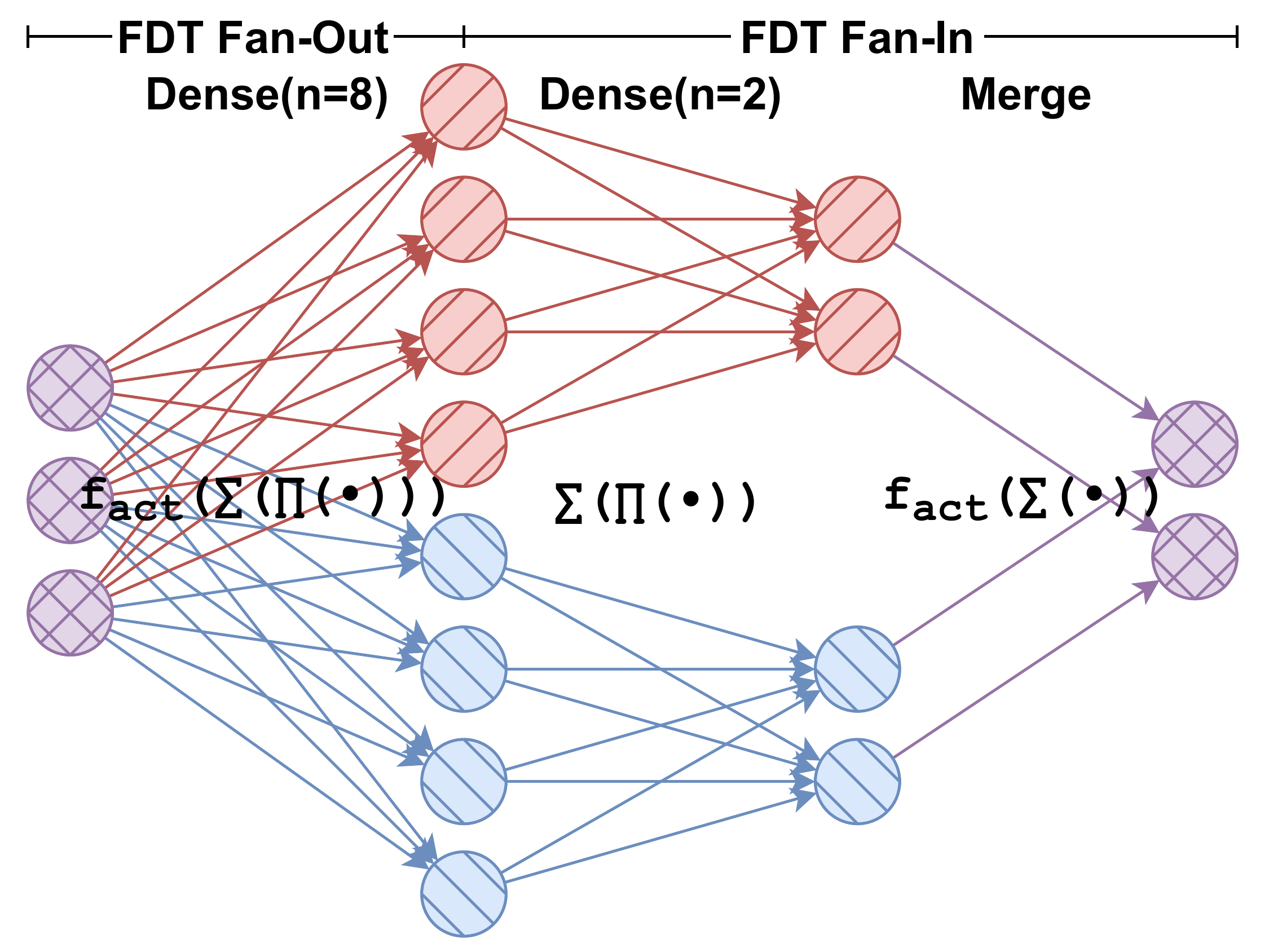}}
    \caption{FDT applied to two dense layers.}
    \label{fig:fdt_fc}
\end{figure}

\Gls{fdt} is first proposed in \cite{stahl2021deeperthings} (named \textit{Fused Layer Partitioning}) for partitioning \gls{dnn} weights of fully connected and convolutional layers that have a large number of weights.
Our work is the first that applies it for the optimization of working memory, i.e. RAM, where the original work targeted the static parameters, i.e. ROM.

The primary goal of fused tiling for memory optimization is the splitting of large intermediate tensor buffers so that their partitions can be computed independently with reduced memory demand.
As shown in Fig.~\ref{fig:ffmt_vs_fdt} on the right, \gls{fdt} does this in the depthwise dimension instead of along the feature maps as with \gls{ffmt}.
Switching to the depthwise dimension avoids any overlap in the intermediate buffer and makes the method independent of the kernel size because the two dimensional convolutions are not split.
However, it requires that the input and output buffers are fully available to every partition, because every single output feature map is the result of summing all input feature maps after applying a convolutional filter.
Fig.~\ref{fig:fdt_fc} helps explain this concept with two consecutive dense (also called fully connected) layers tiled into two partitions.
Part of the output neurons of the first layer (\textit{FDT Fan-Out}) are computed in each partition using all input neurons.
For the second layer (\textit{FDT Fan-In)}, the output neurons can only be computed partially, because not all input neurons are available to every partition.
However, since a dense operation is a sum of products, all partial values of all partitions can be recombined by summing them element-wise and applying the activation function in a new appended \textit{Merge} operation.
Since activation functions are nonlinear, this imposes a limit of two \gls{fdt}-partitioned operations for each tiled sequence.
In \gls{ffmt} there is no inherent limit to the number of consecutive convolutions until the overlap becomes too large to achieve memory savings or the run time overhead becomes impractical.
We call such a tiled sequence \textit{path} and they may contain other operations interleaved with the \gls{ffmt}/\gls{fdt} ones.
For example, element-wise or pooling operations can be inserted, because they do not introduce cross-dependencies between partitions.
\Gls{ffmt} requires spatial locality, while \gls{fdt} can be applied to a wider range of operations where all output elements depend on all input elements as long as there is no interdependence among the output elements.
Examples of operations that can only be tiled by \gls{fdt} are dense operations and pairs of embedding lookup (e.g. TensorFlow \texttt{gather} function) and axis reduction (e.g. by taking the mean).

\section{Automated Tiling Exploration}

\begin{figure}[bt]
    \centerline{\includegraphics[width=0.48\textwidth]{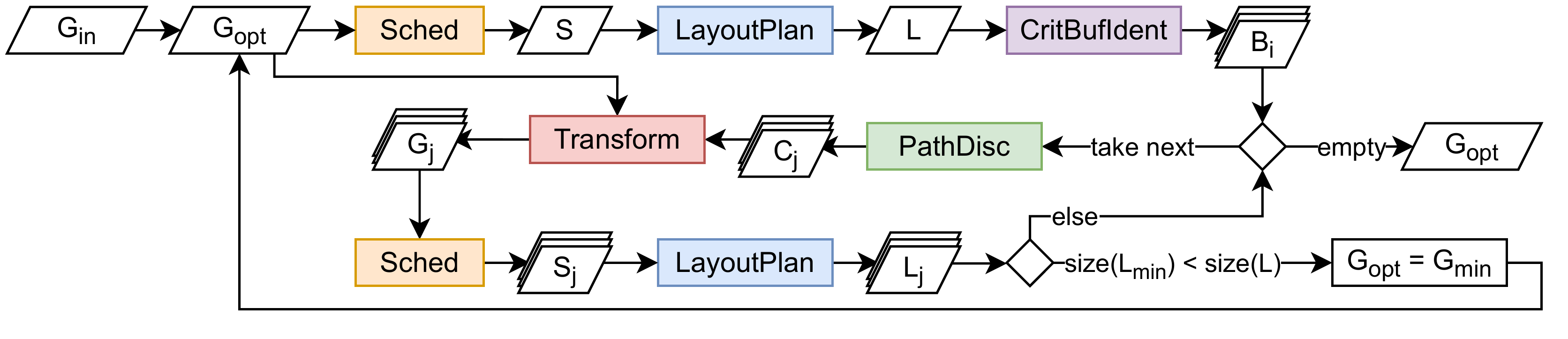}}
    \caption{Automated Tiling Exploration Flow.}
    \label{fig:moiopt_flow}
\end{figure}

It is not meaningful to demonstrate the theoretical memory usage of fused tiling methods in isolation, because the practical memory usage is heavily affected by the entire end-to-end deployment flow with the interdependent problems of tiling configuration, operation scheduling and layout planning.
Each of these problems will be addressed in this section. 
The entire automated tiling exploration flow is shown in Fig.~\ref{fig:moiopt_flow}.
Firstly, the operations of \gls{dnn} graph $G_{in}$ are scheduled in a memory-optimized order $S$.
After the schedule has been fixed, all required intermediate buffers are placed into a linear memory space such that the total required peak memory is minimal.
From the resulting memory layout $L$ a list is extracted that consists of intermediate buffer candidates $B_i$ which may reduce the total memory usage if they were to be tiled.
These $B_i$ are passed to the path discovery in descending order by their size.
The path discovery step identifies tiling configuration candidates $C_j$ for the first buffer candidate.
All configuration candidates are passed to the actual graph transformation pass that applies tiling on the \gls{dnn} graph to produce graph candidates $G_j$.
These are again evaluated by performing scheduling and memory layout planning.
If the memory size of the smallest found layout $L_{min}$ is smaller than the current layout $L$, the corresponding tiling configuration did improve memory usage and the currently best graph candidate $G_{opt}$ is updated.
If no configuration could be found that reduces the memory usage, the next buffer candidate is tested.
The newly generated tiled \gls{dnn} graph $G_{opt}$ is evaluated iteratively.
The flow terminates when no buffer candidate $B_i$ produces a tiling configuration that reduces the layout size further.
In the following, we describe each step in detail, starting with the scheduling and layout planning as these are the prerequisites for path discovery.

\subsection{Memory-aware Scheduling}

For many \glspl{dnn}, scheduling is trivial because their graphs do not contain any branches.
The operation nodes are scheduled in the order as they are located on the single path of the graph.
However, with tiling, parallel paths are introduced in the \gls{dnn} graph and different schedules become possible that determine the lifetime of the intermediate buffers and hence, peak memory.
While optimal memory-aware scheduling has been achieved before in~\cite{ahn2020ordering}, tiled graphs with large number of partitions and many split operations can quickly cause unmanageable run times.
Tiled \glspl{dnn} resemble \textit{series-parallel graphs (SP-graphs)}, i.e. graphs that only comprise of series and parallel compositions of other SP-graphs and the base case of a single node.
Optimal memory-aware scheduling of SP-graphs has been solved with a polynomial-time algorithm by~\cite{kayaaslan2018scheduling} based on~\cite{liu1987application}.
We implemented this algorithm and adjusted the task model to match that of DNN inference.
In contrast to typical task models in distributed computing, the output of an operation can be used by all subsequent operations without distinct buffers for each edge.
For non-SP-graphs, we formulated an \gls{milp}, because we deemed it easier than the method by~\cite{ahn2020ordering}.
If the SP-graph algorithm (it is still $\mathcal{O}(n^3)$) times out, we use a simple heuristic based on \textit{hill-valley segments} introduced in~\cite{liu1987application}, but compromising optimality for trivial run time complexity.
For each parallel path, the heuristic determines the node $N_{i,max}$ with the maximum memory usage and the node $N_{i,min}$ with the minimum memory usage which is also a descendant of $N_{i,max}$.
The paths are now scheduled in their descending order of $N_{i,diff} = N_{i,max} - N_{i,min}$ and used as-is, instead of merging them as in the optimal algorithm.

\subsection{Memory Layout Planning}

After the optimal schedule has been determined, all intermediate buffers of the DNN graph have to be mapped to concrete memory locations.
This is a nontrivial task because buffers can overlap in memory, as long as they are not live at the same time.
The \gls{dnn} graph describes the dependencies between buffers and operations, and the schedule indicates in what order these operations are executed.
Together, these two determine the exact lifetime and, therefore, conflicts that exist between any buffers.
The following \gls{milp} is formulated to perform optimal memory layout planning.
\begin{eqnarray}
  && min_{\mathbf{e}} \quad max_i(e_i)  \label{eq:mip_obj} \\
  s.t. && e_i \geq s_i  \quad e_i \in \mathbb{N}  \quad \forall_{i=1...N}  \label{eq:mip_e} \\
       && e_u - s_u \geq e_v  \lor  e_v - s_v \geq e_u  \nonumber \\
       && \quad \quad (u,v) \in c_j \quad \forall_{j=1...C}  \label{eq:mip_c}
\end{eqnarray}

The $i$-th of a total of $N$ buffers has the ending offset $e_i$ and the size $s_i$. The $j$-th of a total of $C$ conflicts is described by $c_j$ and contains the indices $u$ and $v$ that refer to the buffer list.
The objective function (\ref{eq:mip_obj}) minimizes the largest ending offset of all buffers which is equal to the peak memory usage of all mapped buffers.
The constraint (\ref{eq:mip_e}) ensures that all buffers can only start after the address zero.
Finally, the constraint (\ref{eq:mip_c}) ensures that there are no address overlaps in the list of conflicting buffers.
The nonlinear disjunctions are modeled with the \textit{Big M Method}. 
The final offsets of each buffer are obtained trivially by $e_i - s_i$.

\subsection{Block-based Path Discovery}

Path discovery has the goal of proposing optimized fused tiling configurations that dictate where and how DNN operations are tiled.
The process starts at a buffer that should be split into multiple partitions, called the \textit{critical} buffer.
It then walks the DNN graph up and down to find suitable split and merge points.
After memory-aware scheduling and memory layout planning, the critical buffers are identified by selecting buffers from the memory layout that would reduce the total layout size if their size were to be reduced.
This is achieved by checking whether a buffer is the sole one responsible for the final layout size.
In our work, the input or output buffers of the model cannot be tiled because they are assumed to be written and read as a whole by the application.
The method can be adapted easily if this were not the case.
All critical buffers are considered for path discovery, but the largest ones are checked first.


\begin{figure}[tb]
    \centerline{\includegraphics[width=0.48\textwidth]{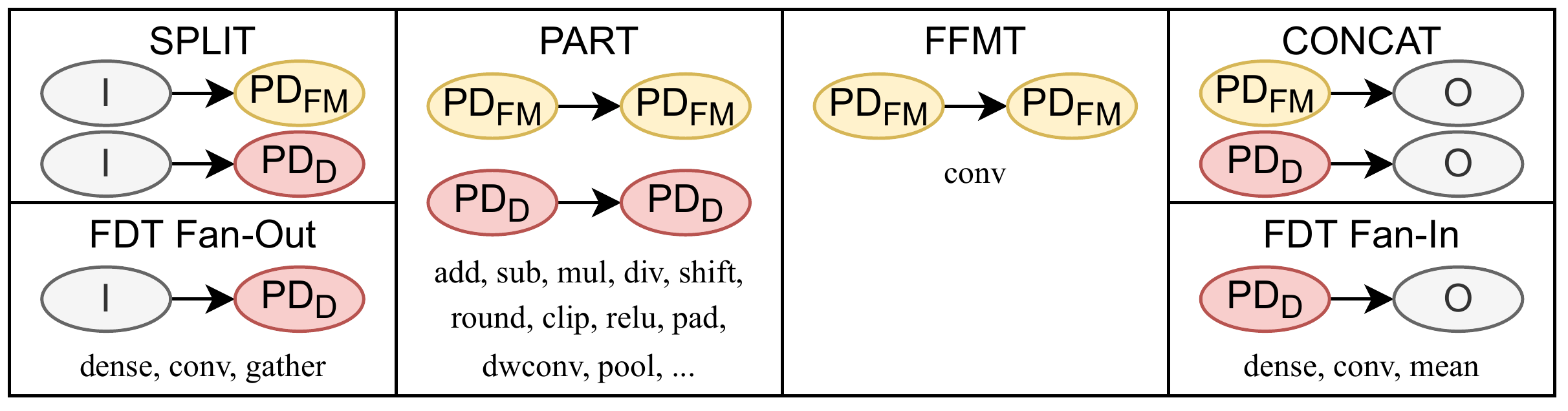}}
    \caption{Path discovery blocks with supported operations.}
    \label{fig:pathdisc_types}
\end{figure}


The blocks of our block-based path discovery along with their supported operations are shown in Fig.~\ref{fig:pathdisc_types}.
The input I is the start of any path and can either be split explicitly by a \textit{SPLIT} operation or implicitly through an \textit{FDT Fan-Out} operation.
\textit{SPLIT} may produce depthwise partitioned (PD\textsubscript{D}) or feature map partitioned (PD\textsubscript{FM}) values.
Partitioned operations (\textit{PART}) compute part of the output values by using their respective input values and are compatible with any partitioned values.
The concatenation operation (\textit{CONCAT}) concatenates multiple partitioned values back into the original non-partitioned buffer O.
The same can be realized implicitly with the \textit{FDT Fan-In} which includes the second partitioned operation of FDT along with the final merge operation as discussed in Section~\ref{sec:fdt}.
\textit{FFMT} represents an operation that is split with \gls{ffmt} and is only applicable to convolutional operations.
\textit{FDT Fan-Out}, \textit{PART}, \textit{FDT Fan-In} and \textit{FFMT} replace their original operation with the tiled variant, while \textit{SPLIT} and \textit{CONCAT} are additionally inserted operations to build a valid path.

\begin{figure}[tb]
    \centerline{\includegraphics[width=0.48\textwidth]{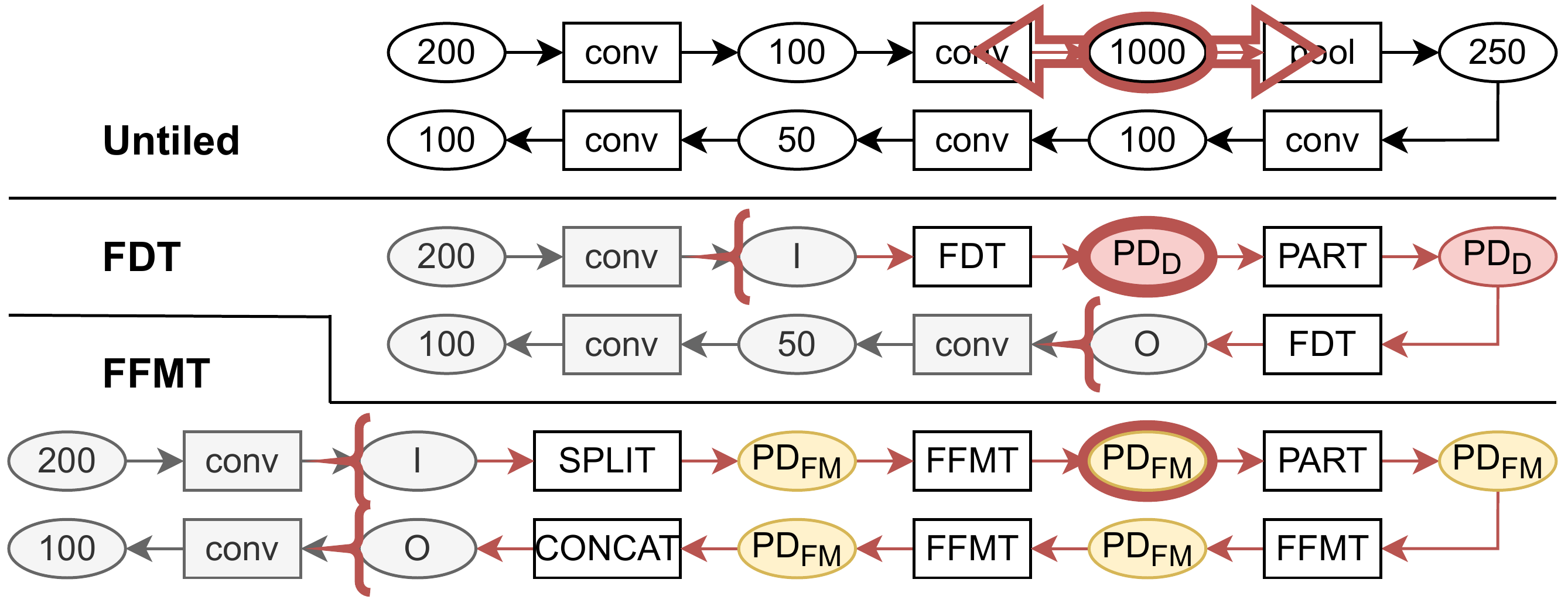}}
    \caption{Path discovery applying FDT and FFMT.}
    \label{fig:pathdisc_example}
\end{figure}

Fig.~\ref{fig:pathdisc_example} shows an example DNN at the top with a highlighted critical buffer.
At the critical buffer, multiple candidate paths are proposed for type PD\textsubscript{D} or PD\textsubscript{FM} if possible.
One proposal is created for each number of partitions $N \in \{2,...,25\}$ with the upper limit chosen to reduce overheads while observing that higher limits rarely provide additional memory savings.
For \textit{FFMT}, quadratic two-dimensional tiling configurations are added as $N \in \{2x2, 3x3, 4x4, 5x5\}$.
Next, the path is discovered starting from the critical buffer in both directions where any compatible block can be chosen.
Whenever the \textit{FDT Fan-In} method is used, one version of the path without \textit{FDT Fan-In} is kept, because a \textit{CONCAT} could require less memory than continuing with partial values.
Whenever an \textit{FFMT}-partitioned operation that has overlap is encountered, one version that stops before that operation is kept and finalized with \textit{SPLIT} or \textit{CONCAT}.
This is done because overlaps that become too large may cause inferior paths compared to shorter ones.
The discovery has to stop at any operation that is incompatible with fused tiling (e.g. softmax, slice, concat).
For each of the proposed path candidates, the operation before the critical buffer with the lowest input buffer size is selected as start of the path and the operation after the critical buffer with the lowest output buffer size is selected as end of the path.
If no such operation could be determined before and after the critical buffer, the path is discarded and if no valid paths are left, the discovery fails.
The second and third graphs in Fig.~\ref{fig:pathdisc_example} show the longest paths of the example for FDT and FFMT respectively.
Note that initially the FFMT path included the outermost convolutions, but since their input/output buffer is larger than the one before, the path terminals are selected as shown.
In the final step, path discovery determines the path that is expected to cause the lowest memory usage.
As mentioned in the overview, this is done by evaluating the memory size with memory-aware scheduling and memory layout planning.
The best configuration is the one with the lowest memory size.

\subsection{Automated Graph Transformation}

Once the best path configuration has been determined, it is applied by transforming the \gls{dnn} graph with the given parameters.
At the start of the split path, either an explicit or implicit split has to be realized.
For an explicit split, a new operation has to be inserted that slices the input into partitions according to the tiling configuration.
An implicit split is implemented by replicating the convolutional or dense layer by the number of partitions and splitting their weight dimension that is responsible for producing outputs.
Any following operations are also replicated on each partition and need their parameters changed to match their new input dimensions.
For example, a bias addition no longer adds its original constants, but only the ones corresponding to the respective partition.
Another example are padding operations where their padding needs to be eliminated at split boundaries to preserve the original \gls{dnn} behavior.
Depthwise convolutions can be split trivially along the channel dimension as tiling method \textit{PART}, since every output channel only depends on its respective input channel.
The associated filter weights must still be split accordingly.
The exact splitting logic for every operator has to be determined on a case-by-case basis.
However, it is possible to define categories with similar splitting logic.
\textit{FDT Fan-In} operations are split equivalently to \textit{FDT Fan-Out} ones, just that the input channel dimension of the weight tensor is split.
Care has to be taken to prohibit automatic fusing of the last operations on the split paths with the \textit{CONCAT} or \textit{FDT Fan-In} operation, because that would lead to keeping their inputs alive on multiple split paths.
After all transformations have been applied to the graph, the flow goes back to scheduling it as shown in Fig.~\ref{fig:moiopt_flow}.

\subsection{Implementation}
The complete end-to-end flow for comparing \gls{fdt} to \gls{ffmt} has been implemented in \textit{Apache TVM}~\cite{chen2018tvm}.
TVM is a state-of-the-art machine learning compiler that takes \glspl{dnn} and converts them into its own \gls{ir}.
This \gls{ir} is used to optimize the \gls{dnn} through compiler transformations that are aware of the machine learning domain.
Finally, various backends are able to produce output for different deployment targets like GPUs or microcontrollers.
We chose TVM because its \gls{ir} is very suitable for implementing complex transformation passes.
To achieve competitive results compared to widely-used frameworks like \textit{TensorFlow Lite for Microcontrollers}, we chose the \textit{Ahead-of-Time (AoT) TVM backend} that generates static code that is able to run the \gls{dnn} inference without the full TVM run-time libraries.
In TVM, many \gls{dnn} operations are fused to eliminate intermediate buffers entirely.
For example, a convolution with bias addition and activation function is carried out by adding the bias and applying the activation function while calculating each individual convolution output value.
All intermediate buffers between such fused operations do not contribute to the peak memory usage.
Therefore, when analyzing a \gls{dnn} for critical buffers, only the buffers of non-fused operations are taken into consideration.
However, during path discovery, all fused operations are transformed into their fine-grained operations because they may contain operations that are suitable as terminals of the split path.
While this may have an effect on inference latency through increased number of memory accesses, the goal of the optimization is having small buffers at the path terminals, so that those possible extra accesses will never dominate other accesses.
After the graph transformation step, operations are fused again at every possible opportunity.

\section{Results}

From a wide range of models, the following subset was identified that benefits from fused tiling.
All models are quantized to 8 bits.

\begin{enumerate}
    \item \textit{Keyword Spotting (KWS)}: Detection of keywords from audio. Part of the MLPerf Tiny benchmark~\cite{banbury2021mlperf}.
    \item \textit{Text Sentiment Analysis (TXT)}: \cite{text_class,maas-EtAl:2011:ACL-HLT2011}.
    \item \textit{Magic Wand (MW)}: TinyML gesture recognition with an accelerometer~\cite{david2020tensorflow}.
    \item \textit{PoseNet (POS)}: Pose estimation~\cite{papandreou2018personlab}.
    \item \textit{MobileNet V2 SSDLite (SSD)}: COCO classifier~\cite{sandler2018mobilenetv2}.
    \item \textit{Cifar10 classifier (CIF)}: Own CNN~\cite{krizhevsky2009learning}.
    \item \textit{Radar Gesture Recognition (RAD)}: Own TinyML CNN for gesture recognition with a radar sensor.
\end{enumerate}

The target architecture for all experiments was RISC-V in the \textit{RV32GC} configuration.
The GNU toolchain at version 11.1.0 was used with the optimization flag set to \texttt{-Os} and options to prune all unused code and data.
The RAM and ROM usage is determined from the section sizes in the compiled binary.
The run time is estimated by statically determining the number of multiply-accumulate (MAC) operations required in the final optimized DNN graph.
This gives a good estimate because the computational cost of DNNs is dominated by matrix multiplications and therefore MACs~\cite{tsai2018recent}.
This is not equivalent to the run time after deployment, but is sufficient for a relative comparison.
Dynamic instruction counts were also gathered on an instruction set simulator, but they showed high sensitivity to TVM's automatic kernel generation, rather than the chosen tiling configurations.
This could be explained by TVM's lack of operator schedules that are optimized for RISC-V.
The \glspl{milp} were implemented in ORTools 9.3~\cite{ortools} using the Gurobi 9.1.2 solver~\cite{gurobi}.

\subsection{Automated Tiling Exploration}

Our optimal memory layout planning algorithm was compared to the best-performing heuristic approach in TVM that uses hill-climbing and simulated annealing.
The heuristic finds the optimum for most models, but in one case (the TXT model) we achieved a memory reduction of 16.8\% compared to the heuristic.

Our MILP memory-aware scheduling solution is optimal, as defined by its cost function.
The work in \cite{ahn2020ordering} reports a run time of 37.9 seconds for the SwiftNet model~\cite{cheng2019swiftnet}.
When running our MILP scheduling with the same SwiftNet model, we measured a run time of 37 seconds.
While not being able to directly compare these numbers due to different used machines, our result on an \textit{AMD Ryzen 9 3900X} processor shows comparable performance.

Our path discovery is able to traverse a large variety of models and selects the optimal solution within its search space.
This search space ranges from zero to hundreds depending on the critical buffer dimensions and operations used to create a path.
Further factors are the variants with early path stops and the iterative application of tiling.
The innermost operations of graph transformation, scheduling and layout planning have to be executed that number of times.
For the evaluated models, the entire flow has a run time of 3 minutes for the RAD model (38 tiling configurations) up to an hour for the POS model (172 tiling configurations).
\cite{arm2021cascading,minakova2020buffer,minakova2022memory,lin2021mcunetv2,colleman2021high} do not provide flow run times.
The work of \cite{cipolletta2021dataflow} reports 82 to 375 seconds for searching nine configurations, while still having to manually select the number of partitions and their axes.
This shows that the implemented flow runs efficiently and, in contrast to existing work, requires no manual choice for the tiling configuration.

\subsection{Fused Depthwise Tiling}

\begin{table}[tb]
    \caption{Memory reduction of FDT compared to FFMT}
    \centering
    \scalebox{0.74}{
    \begin{tabular}{|c|c|cc|cc||c|cc|cc|}
        \hline
         & \multicolumn{3}{c|}{Mem [kB]} & \multicolumn{2}{c||}{[\%]} & \multicolumn{3}{c|}{MACs [1 million]} & \multicolumn{2}{c|}{[\%]} \\
        \hline
        \rotatebox{90}{Model} & \rotatebox{90}{Untiled } & \rotatebox{90}{FFMT} & \rotatebox{90}{FDT} &  \rotmulticell{FFMT \\ Savings} & \rotmulticell{FDT \\ Savings} & \rotatebox{90}{Untiled} & \rotatebox{90}{FFMT} & \rotatebox{90}{FDT} & \rotmulticell{FFMT \\ Overhead} & \rotmulticell{FDT \\ Overhead } \\
        \hline
        KWS   & 65.6  & 65.6  & 53.7  & 0.0 & 18.1  & 2.66 & 2.66 & 2.66 & 0.0   & 0.0  \\
        TXT   & 18.6  & 18.6  & 4.43  & 0.0 & 76.2  & 0.00 & 0.00 & 0.00 & 0.0   & 0.0  \\
        MW    & 17.6  & 7.04  & 11.3  & 60.9 & 35.5 & 0.06 & 0.06 & 0.06 & 0.0  & 0.0  \\
        POS  & 9.35k  & 5.11k & 8.94k & 45.3 & 4.4 & 837  & 1215 & 837  &  45.1  & 0.0  \\
        SSD   & 14.3k & 8.66k & 12.2k & 39.4 & 14.6 & 313  & 314 & 313  &  0.2 & 0.0  \\
        CIF &  179   &  76.7  &   170 &  57.1 & 5.0  & 5.52  & 6.02  & 5.52  & 9.0 & 0.0  \\
        RAD &  36.2  &  26.7  &  29.4 &  26.3 & 18.8 & 0.09  & 0.09  & 0.09  &  0.0  & 0.0  \\
        \hline
        \textbf{Avg.} &    &  &       & 32.7  & 24.7  &     &       &       & 7.8  & 0.0  \\
        \hline
    \end{tabular}}
    \label{tab:fdt_vs_ffmt}
\end{table}

The results in Table~\ref{tab:fdt_vs_ffmt} show the working memory (RAM) usage and estimated MAC operations for each untiled network and the improvements by applying \gls{ffmt} or \gls{fdt} individually.
The first two models are only able to be tiled by \gls{fdt}.
In the case of KWS, the critical buffer is involved in a sequence of convolutions that reduce the feature map size down to 1x1, which can not be split by \gls{ffmt}.
The TXT model's critical buffer exists within an embedding lookup followed by a mean axis reduction that can only be tiled by \gls{fdt}.
The remaining models are all CNNs with sufficient feature map sizes such that either method is applicable.
\Gls{fdt} never incurs any run time overhead at the cost of lower memory reduction compared to \gls{ffmt}.
The average memory savings are 32.7\% for FFMT and 24.7\% for FDT with the highest savings achieved for the TXT model with FDT at 76.2\%.
Mostly, the run time overheads of both methods are negligible, but the POS and CIF models with FFMT showed a significant overhead of 45.1\% and 9.0\% because they contain larger chains of fused operations that cause more redundant calculations from overlapping partitions.
In these cases, FDT offers an alternative design point without any overhead, but often reduced memory savings.
For the remaining three models, FDT did not achieve higher memory savings than FFMT and FFMT did not cause a significant run time overhead.
The limitation of FDT is therefore its limited applicability to models in general.
The ROM overheads are not shown because they are negligible with impacts below 1\%.
Although~\cite{stahl2021deeperthings} used FDT and FFMT as well, that work only investigated memory usage without inference, which mainly amounts to ROM usage.

Enhancing an FFMT-only TinyML deployment flow with FDT expands the tiling design space for memory and performance goals.
In the case of a memory-optimized design, the fused tiling method with the highest memory savings can be selected.
In the case of a performance-optimized design, the highest memory savings should be selected with the constraint that the run time overhead may not exceed a certain threshold.
The exploration also found tiling configurations, in which FFMT and \gls{fdt} are applied in conjunction.
However, the results were in the best case as good as the best configuration with a single tiling method.
Still, for possible new models, the combination could also yield benefits.

\section{Conclusion}

In this paper, we apply Fused Depthwise Tiling to \gls{dnn} graphs for memory optimization for the first time.
We built a state-of-the-art, end-to-end deployment flow for its evaluation.
In TinyML scenarios, integrating this new tiling reduces the memory usage of two evaluated models significantly and offered additional design points for two other models that eliminate the run time overhead at reduced memory savings.

\begin{acks}

This work was supported in part by the German Federal Ministry of Education and Research (BMBF) within the project Scale4Edge under contract no. 16ME0131.

\end{acks}

\bibliographystyle{ACM-Reference-Format}
\bibliography{bib}

\end{document}